\pdfoutput=1

\documentclass[11pt]{article}

\usepackage{EMNLP2022}

\usepackage{times}
\usepackage{latexsym}

\usepackage[T1]{fontenc}

\usepackage[utf8]{inputenc}

\usepackage{microtype}

\usepackage{inconsolata}
\usepackage[bottom]{footmisc}

\usepackage{xcolor}
\usepackage{mdframed}

\usepackage{url}
\usepackage{amsmath}
\usepackage{amssymb}
\usepackage{booktabs}
\usepackage{graphicx}
\usepackage{varwidth}
\usepackage{tabularx}
\usepackage{algorithm}
\usepackage[noend]{algpseudocode}

\usepackage[utf8]{inputenc}

\usepackage[textsize=scriptsize,disable]{todonotes}

\DeclareMathOperator*{\argmax}{argmax}

\title{Deductive Additivity for Planning of Natural
Language Proofs}

\author{Zayne Sprague \quad\quad
  Kaj Bostrom \quad\quad
  Swarat Chaudhuri \quad\quad
  Greg Durrett \\
  Department of Computer Science\\
  The University of Texas at Austin\\
  \texttt{\{zaynesprague@, kaj@cs, swarat@cs, gdurrett@cs\}.utexas.edu}
  }

\date{}

\begin{document}
\maketitle
\begin{abstract}

Current natural language systems designed for multi-step claim validation typically operate in two phases: retrieve a set of relevant premise statements using heuristics (planning), then generate novel conclusions from those statements using a large language model (deduction). The planning step often requires expensive Transformer operations and does not scale to arbitrary numbers of premise statements. In this paper, we investigate whether an efficient planning heuristic is possible via embedding spaces compatible with deductive reasoning. Specifically, we evaluate whether embedding spaces exhibit a property we call \textit{deductive additivity}: the sum of premise statement embeddings should be close to embeddings of conclusions based on those premises. We explore multiple sources of off-the-shelf dense embeddings in addition to fine-tuned embeddings from GPT3 and sparse embeddings from BM25. We study embedding models both intrinsically, evaluating whether the property of deductive additivity holds, and extrinsically, using them to assist planning in natural language proof generation. Lastly, we create a dataset, Single-Step Reasoning Contrast (SSRC), to further probe performance on various reasoning types. Our findings suggest that while standard embedding methods frequently embed conclusions near the sums of their premises, they fall short of being effective heuristics and lack the ability to model certain categories of reasoning.

\end{abstract}

\section{Introduction}

One way to justify the truth of a statement is to give an explanation building logically towards that statement based on deduction from shared premises. The ways facts can be combined through reasoning are numerous, including many different modes of deduction like syllogism or modus tollens. This process can be automated with natural language processing, using systems to generate natural language proofs that use evidence to derive a claim through a structured argument. Large language models (LLMs) like GPT4 \citep{openai2023gpt4} have exhibited impressive performance in reasoning tasks. However, these models can still make unsound inferences \cite{ye2022unreliability, zhang2023language, Xue2023RCOTDA}.

\begin{figure}[t!]
    \centering
    
    \includegraphics[width=1.0\linewidth]{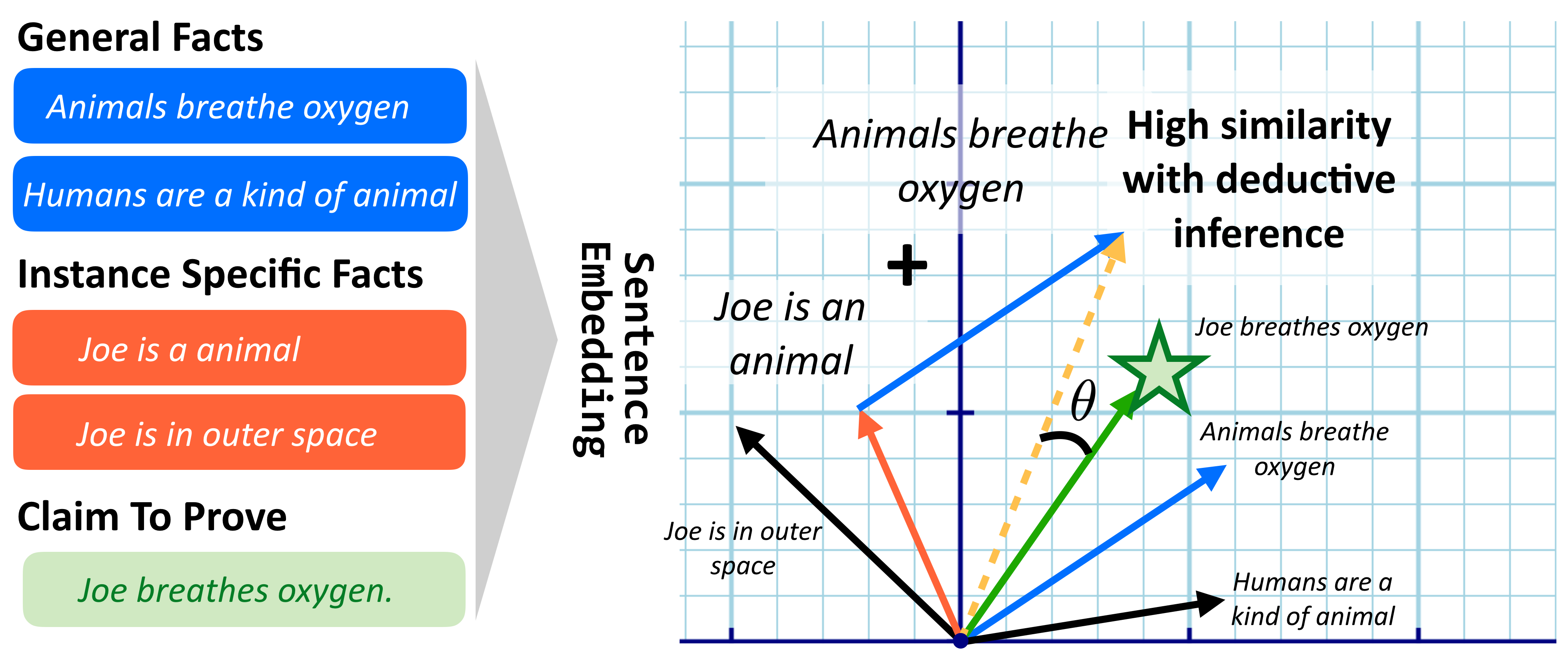}
   
    \caption{ 
    A visualization of an embedding space that has the Deductive Additivity property.  When two facts (blue and red) are added together, their resulting vector (yellow) should have high similarity with the embedding of a statement that logically follows via deduction (green).
    }
    \label{fig:intro_diagram}
\end{figure}

One reason for these errors is that models may fail to plan reasoning effectively. LLMs do not have explicit planning capabilities: they generate conclusions in a way that conflates lexical choice and decisions of what content to generate, and no alternatives are materialized in typical greedy or sampling-based LLM inference. A recent line of work \citep{bostrom-etal-2021-flexible,bostrom-etal-2022-natural,sprague-etal-2022-natural,creswell2023selectioninference} explores how to decouple these stages. However, what is still missing is a scalable method for doing planning in these kinds of natural language reasoning settings: past work involves early-fusion invocation of pre-trained LMs \citep{xiong2021answering} and does not scale to thousands of premises.

This work explores the feasibility of planning the reasoning process directly in a vector space, where combining statements and retrieving similar statements can be efficiently implemented as addition and cosine similarity, respectively. We introduce \textit{deductive additivity (DA)}, a property of an embedding space necessary to enable this planning. A visualization of an embedding space with the deductive additivity property is shown in Figure~\ref{fig:intro_diagram}. Each piece of evidence is embedded into a fixed-size vector, and the combined embeddings of two facts should be close to embeddings of statements that are entailed from those two facts via deduction. This property can help us plan when we are trying to derive a goal statement based on premise statements. New facts that bring us closer to that goal should be explored in the deductive reasoning process, so this vector space provides a natural heuristic: we want to find fact embeddings that, when summed, achieve the highest dot product with the encoding of our goal. Crucially, the vector-based nature of this heuristic facilitates rapid retrieval through efficient search algorithms.

Our experiments test both off-the-shelf embeddings (e.g., SimCSE \cite{gao-etal-2021-simcse}) as well as embeddings that are explicitly tuned for deductive additivity. First, we conduct intrinsic evaluations to see whether embeddings of standard encoders exhibit deductive additivity. We then test how well the method performs as a search heuristic on the natural language proof generation datasets EntailmentBank \cite{dalvi-etal-2021-explaining} and Everyday Norms: Why Not \cite[ENWN]{sprague-etal-2022-natural}. Finally, we create the Single-Step Reasoning Contrast (SSRC) dataset to benchmark each method on how well they model different reasoning categories, like syllogism or modus tollens, and how robust they are to common errors in reasoning, like negation\footnote{Code and data publicly available at  \url{https://github.com/Zayne-sprague/Deductive_Additivity_for_Planning_of_Natural_Language_Proofs}}.

Our main contributions are threefold: (1) We propose a novel method for planning reasoning steps over a collection of facts purely based on vector arithmetic. (2) We show that several embedding methods have promise for deductive additivity but do not fully meet the properties required for planning in natural language deduction scenarios even when explicitly fine-tuned for it. (3) We present a new dataset meant to help diagnose and identify areas where deduction planning methods are underperforming across a range of different reasoning categories.

\section{Problem Description and Motivation}

Here we introduce the problem of proof generation, the system we use to generate proofs and deductive additivity.

\subsection{Problem Setup}

We explore the process of proving a goal statement (or claim) $g$ by generating an entailment tree $T$, given a set of general-purpose facts $X = {x_1,\ ...\ x_n}$ and a collection of instance-specific facts $F = {f_1,\ ...\ f_m}$. Instance-specific facts typically pertain to the context or background of a particular scenario, while general-purpose facts can be applied more broadly. An example can be seen in Figure~\ref{fig:intro_diagram}, where $F$ consists of two statements, ``\emph{Joe is an animal}'' and ``\emph{Joe is in outer space}'', and all other facts belong to $X$. $T$ is a binary-branching tree with its leaves being members of $X$ and $F$ while its non-leaf nodes (which we also call \emph{intermediates}) are new statements generated via deductive reasoning. The root of $T$ must logically entail $g$. We use the entailment models from past work \citep{bostrom-etal-2022-natural,sprague-etal-2022-natural}, which are based on WaNLI \cite{liu-etal-2022-wanli} to make this judgment.

The EntailmentBank dataset \citep{dalvi-etal-2021-explaining} formalizes three variants of this problem setting. The first setting, denoted as Task 1 (T1), provides only the general-purpose facts relevant to the construction of the gold entailment tree, making it the easiest setting as it eliminates the need to sift through irrelevant facts. Task 2 (T2) includes both the relevant facts and lexically similar distractor facts. Task 3 (T3) \citep{dalvi-etal-2021-explaining} includes all facts from a large corpus like Wikipedia as the general-purpose fact set $X$. In all these settings, the task involves iteratively building the entailment tree through deductions until the original goal $g$ is entailed. Our experiments will focus on the T2 setting. \footnote{While the T3 setting offers a large-scale stress test for retrieval-based approaches like ours, we found in practice that a first-stage retrieval (i.e., converting T3 to T2) with BM25 worked well for all datasets considered in this work. Nevertheless, models that scale to large $X$ sets will be useful for future systems tackling more sophisticated problems like automatic fact-checking.}

\subsection{Proof Generation}
\label{sec:proof_search_intro}

We follow past work on these tasks \citep{bostrom-etal-2022-natural,sprague-etal-2022-natural} where the intermediate nodes of the entailment tree are generated from a pre-trained language model. Details on the model are in Appendix \ref{sec:appendix_search_modules}. Specifically, given two premise statements $p_a$ and $p_b$, we assume access to a model $P(d_{ab} \mid p_a, p_b)$ that places a distribution over valid deductions $d$ given the two premises. If the two premises do not combine to yield any meaningful new conclusions, the behavior of this system is not well-defined.

To produce an entailment tree $T$, we follow the proof generation algorithm from \citet{bostrom-etal-2022-natural}; we outline it here and detail all modules of the search algorithm in Appendix~\ref{sec:appendix_search_modules}. We begin with our collection of premises $P = \{X \bigcup F\}$. In EntailmentBank and ENWN, the set $P$ is given per dataset example. From $P$, a heuristic $M$ ranks pairs of premises as to how useful their deduction will be in proving the claim $g$ (also given per example). We denote a single ranked premise pair as a step in the search, and we term the current collection of steps at any moment in the search as the search fringe.

A deductive step model, $S$, pops the highest-ranked step (according to $M$) from the fringe and generates a set of deductions.\footnote{To thoroughly explore the space of all plausible deductions, we sample $k$ generations each time ($k = 5$ in all our experiments).} These deductions are validated and added back to the pool of premises $P$, where the heuristic will rank all potential pairs of the new set of deductions with all other previous premises to create new steps in the search fringe. This process is repeated until the $maxSteps$ limit is reached or the fringe has been exhausted. 

Our work focuses on investigating if the heuristics used during the search can leverage embedding spaces that exhibit deductive additivity.

\subsection{Deductive Additivity}
\label{sec:additive_reasoning_intro}

Recall that $d_{ab}$ represents a valid conclusion from a pair of premises $p_a$ and $p_b$. Our heuristics are based on an embedding function $E: \Sigma^* \rightarrow \mathbb{R}^n$, embedding a sentence into $n$-dimensional space. We represent the sum of the embedded premises as the deductive trajectory embedding $\mathbf{e}^{'}_{a+b} = E(p_a) + E(p_b)$, where $\mathbf{e}^{'}$ signifies embeddings produced through arithmetic operations rather than the encoder $E$. An encoder $E$ generates an embedding space exhibiting the property of deductive additivity if the deductive trajectory embedding has a higher cosine similarity with their embedded conclusion than any other statement, $x$, not entailed by the premises via deduction, denoted as $p_a, p_b \nrightarrow x$. That is, we want
\begin{equation}
\cos(\mathbf{e}^{'}_{a+b}, E(d_{ab})) > \cos(\mathbf{e}^{'}_{a+b}, E(x))
\label{eq:additive_reasoning_def}
\end{equation}

\noindent When the condition in Equation~\ref{eq:additive_reasoning_def} holds, the embedding space is capable of representing logical relationships strictly in their vectors and can be expressed through simple arithmetic operations such as addition.

\subsection{Tuning for Deductive Additivity}
\label{sec:additive_reasoning_training}

Any sentence embedding method can be evaluated for whether or not it exhibits deductive additivity. However, we additionally describe a method for fine-tuning an embedding model to have this property.

We use EntailmentBank to obtain a collection of premise deduction triplets $D = \{p_a,\ p_b,\ d_{ab}\}$. Subsequently, we use a loss function to push the encoded representations of the premises closer to that of the deduction \cite{pmlr-v119-chen20j, gao-etal-2021-simcse}.

\begin{equation}
l_{ab} = - \log \frac{\mathrm{exp}(\textbf{e}^{'}_{a+b} \cdot E(d_{ab})/\tau)}{\sum^N_{i=1} \mathrm{exp}(\textbf{e}^{'}_{a+b} \cdot E(d_i) / \tau)} 
\label{eq:additive_loss}
\end{equation}

where $N$ represents the batch size. Most deductions $d_i$ will not entail the deduction $d_{ab}$, so they serve as suitable negatives from the perspective of Equation~\ref{eq:additive_reasoning_def}. 

For training, we employ temperature scaling in the contrastive loss in Equation~\ref{eq:additive_loss}. Previous work has found that contrastive learning benefits from having large batch sizes, more in-batch negatives, and hard negatives \cite{he2020momentum, karpukhin-etal-2020-dense, DBLP:journals/corr/abs-2003-04297, radford2021learning, xiong2021answering}. To take advantage of hard in-batch negatives, we leverage the tree structures in our training data (EntailmentBank). Specifically, each batch in our training loop contains all the intermediate labeled steps for an entailment tree in EntailmentBank, covering multiple trees. We discover that triplets from the same tree serve as suitable proxies for hard negatives in our contrastive learning process, allowing us to bypass the need for hard negative mining. Our batches include 100 trees, as many as we could fit onto our GPU, which equates to 200-300 triplets in a batch. We found that increasing the batch size led to better performance. We implement our method with the PyTorch Metric learning library \cite{Musgrave2020PyTorchML}.

Following each epoch of training, we assess the encoder's performance by our second intrinsic evaluation, Ranking Gold Steps. We use the EntailmentBank T2 development set for checking when to stop training the encoder. 

\subsection{Caching}

Certain heuristics used in proof generation algorithms, such as the one we construct using deductive additivity, can cache the encodings of the initial evidence pool $X$. This offers significant time savings in completing the first step of a search procedure (where a non-cached method would need to set up and rank the pairs for the initial set). However, any subsequent deductions will need to be encoded since they cannot be precomputed and cached. We also found the time savings to be relatively limited in the $T1$ and $T2$ settings since $n$ is relatively small, so we do not expand on this capability further.

\section{Heuristics and Datasets}

To measure the performance of using deductive additivity as a proof generation heuristic, we explore five heuristics and three datasets.

\subsection{Baseline Heuristics}

We consider two baseline heuristics for ranking and retrieving relevant statements: BM25, a sparse retrieval method, and the original heuristic from previous work, SCSearch, which employs an early-fusion premise ranker model.

\paragraph{BM25} BM25 \citep{robertson1995okapi} matches items in an index with a query via sparse vector representations, capturing lexical overlap but not deeper semantic similarity. In the proof generation search procedure, we index all concatenations of strings in each step (two premises, generated deductions, or one of both), then retrieve the best step based on the goal. 

\paragraph{SCSearch} Past work \citep{bostrom-etal-2022-natural} has used heuristics with a substantially different structure. These heuristics use language models like DeBERTa to score premise pairs conditioned on a claim. Specifically, these models are of the form $\mathbf{w}^\top E(p_1,p_2,g)$; they encode $p_1$, $p_2$, and $g$ jointly with an encoder model. A linear layer $\mathbf{w}$ is then used to predict a logit value used for ranking. These models are trained as binary classifiers on EntailmentBank by selecting positive examples of premise pairs that eventually lead to $g$ and negative examples of unrelated premise pairs. This allows the language model to determine if the immediate deduction would be beneficial towards deducing the claim that it is conditioning on. It also allows the language model to see the claim and premise pairs in context and model interactions between them. Because these methods use Transformers to score the premise pair and can model nonlinear interactions between the premises, these models are strictly more expressive than vector-based heuristics.

\subsection{Embedding-based Heuristics}

To test if embeddings with deductive additivity can be useful in proof generation, we employ three different heuristics that all use deductive additivity but with different encoders to compare different embedding spaces. A deductive additivity heuristic will, for each step, encode any new deductions from the previous step and then sum all the pairs to create deductive representations $\mathbf{e}^{'}_{d}$ for hypothetical deduced pairs. We then compute the cosine similarity of each $\mathbf{e}^{'}_{d}$ with $\mathbf{e}_g$ (the goal embedding), which is used as a score to select the next step $S_i = \underset{d}{\mathrm{argmax}}\ \mathrm{cos}(\mathbf{e}^{'}_{d}, \mathbf{e}_g)$. 

We consider the deductive additivity heuristic under three different encoders: SimCSE and GPT3 are used to test off-the-self sentence encoders for deductive additivity, and finally, we fine-tune GPT3 explicitly for deductive additivity.

\paragraph{SimCSE} SimCSE \citep{gao-etal-2021-simcse} is an encoder that produces sentence embeddings optimized using a contrastive objective.\footnote{Note that this contrastive objective is different from ours. Training for SimCSE was performed on natural language inference (NLI) examples from MNLI and SNLI datasets. From the perspective of data assumptions, we place it in the ``fine-tuned'' category; although it hasn't been trained on EntailmentBank data explicitly, it uses related entailment data.} We test to see if this encoder produces an embedding space where deductive additivity holds.

\paragraph{GPT3} We use OpenAI's embedding endpoint to create sentence embeddings using the Ada model \citep{NEURIPS2020_1457c0d6}. We test to see if this encoder produces an embedding space where deductive additivity holds as well.

\paragraph{GPT3-tuned} We combine OpenAI's embedding endpoint with three additional dense layers using the GLU activation function with residual connections between each layer. We then fine-tune these three layers using the EntailmentBank T1 dataset as described in Section~\ref{sec:additive_reasoning_training}.

\subsection{Datasets}
\label{sec:datasets}

\paragraph{EntailmentBank (EB)} This dataset comprises annotated entailment trees for textbook-based science facts \citep{dalvi-etal-2021-explaining}. We used this dataset for training the majority of our models in a T1 setting. We evaluate the models on the test slice of entailment trees for the T2 task setting.

Each example in EB contains a set of premises, $P$, and a claim $g$ that we are trying to prove given $P$. To prove $g$, the system has to produce a series of deductions by combining two premises from the set $P$, then combining intermediate deductions and the premises in $P$ until the claim is proven. Whether it is proven is determined via an entailment model scoring $g$ above a certain threshold from some generated conclusion following previous work \cite{sprague-etal-2022-natural, bostrom-etal-2022-natural} and detailed further in Appendix \ref{sec:appendix_search_modules}. Planning heuristics must determine which premise-premise or premise-deduction pairs are most likely to help in proving the claim, as the set of pairwise premises and intermediate deductions can be large. 
 
In the T2 setting, the number of premises $n$ is fairly small; $n < 30$ for most examples. There are usually only 3 to 5 deductions involved to produce the annotated entailment tree. We allow for a total of 10 steps ($maxStep$), and for each step, we allow for five generations to be sampled ($k$).

\paragraph{Everyday Norms: Why Not (ENWN)} ENWN \citep{sprague-etal-2022-natural} contains annotated entailment trees for common everday scenarios. Structurally, ENWN resembles EntailmentBank but with a different domain of reasoning and a larger number of required deductive steps on average (4.71 to 4.26). ENWN aims to combine common social rules deductively to determine whether a person should perform a particular action (usually something they should not do). ENWN currently does not have a T2 or T3 setting. 

\subsection{Single-Step Reasoning Contrast Dataset} Both EntailmentBank and ENWN test a subset of logical inference types but do not necessarily have broad coverage. For example, EntailmentBank has very few examples involving negation, despite this being a very important phenomenon to model in practice. We want to test whether our embedding methods can handle a wider range of cases.

We construct a new dataset that examines common forms of logical reasoning\footnote{We initially employed ChatGPT for annotating examples in EntailmentBank and ENWN. However, it did not yield consistent labels, signaling an opportunity for further exploration in future research. Instead, we adopted a different approach, generating a selection of widely-used labels that we subsequently employed as the reasoning categories within the SSRC dataset.} via synthesized examples. We consider fourteen categories: Analogy, Categorical Syllogism, Causal reasoning, Classification, Comparison, Composition, Division, Modus Ponens, Modus Tollens, Definition, Temporal Logic, Propositional Logic, Quantificational Logic, and Spatial Relationship. For each category, we use GPT-3.5 to generate ten examples of deductions given two premises using the corresponding reasoning category.

For every example deduction, we prompt GPT 3.5 further to perturb the premises in four ways creating additional examples of incorrect deductions. For each perturbation, we create three examples where one or both premises have been \emph{negated}, three examples where one or both premises are a \emph{false premise}, fifteen examples where one or both premises are an \emph{irrelevant fact}, and three examples where one or both premises have an \emph{incorrect quantifier} (usually meaning that ``some'', ``all'', or ``none'' has been prepended to the premise). Examples from the dataset from different reasoning categories and perturbation types are shown in Section~\ref{sec:ssrc_examples} of the Appendix in Table~\ref{tab:rb_examples}. Prompts to create examples and perturb the examples can be found in Appendix \ref{sec:appendix_ssrc_prompts}.

\begin{figure}[t!]
    \centering

    \includegraphics[width=1.0\linewidth]{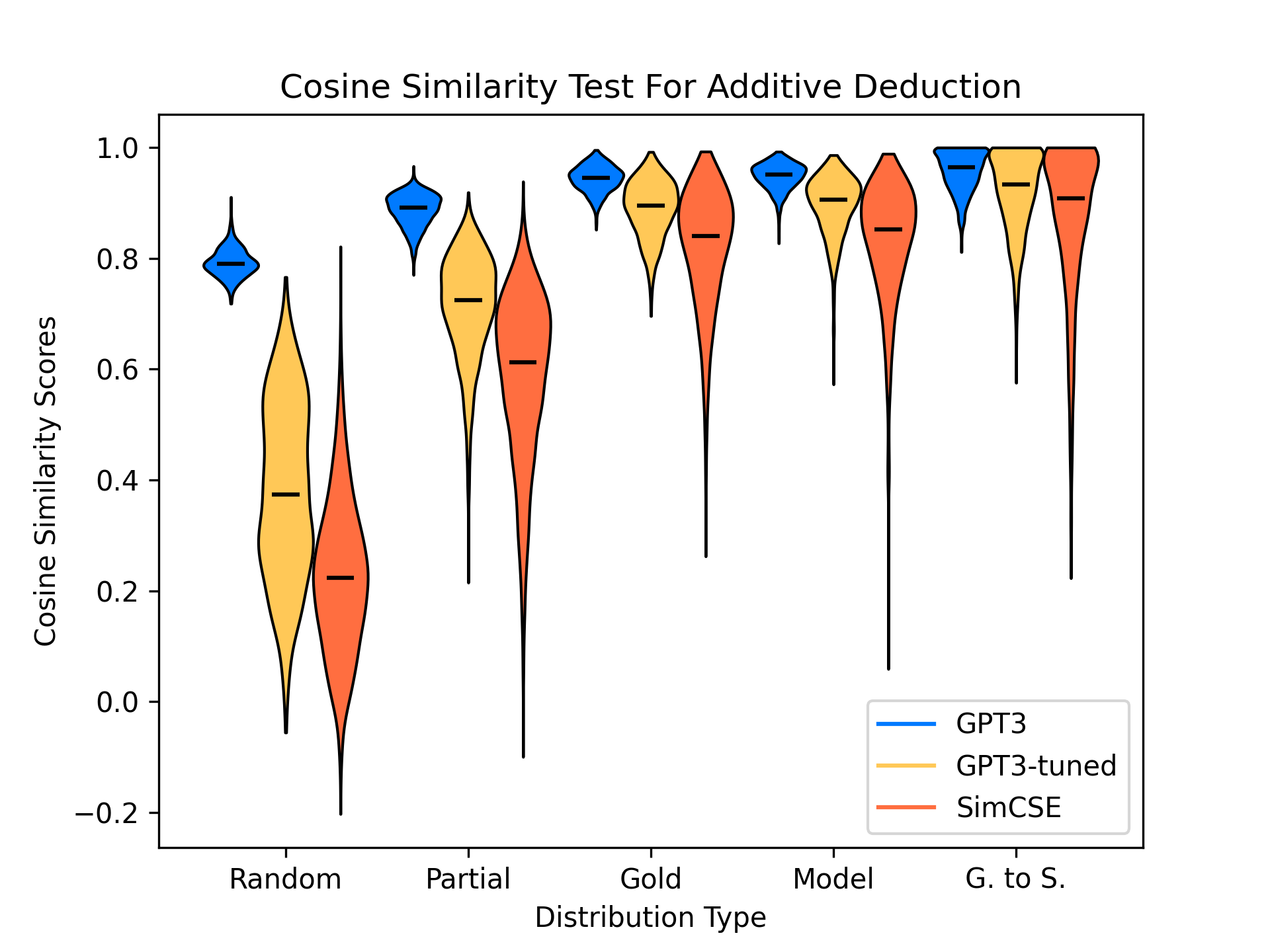}
   
    \caption{Distribution of cosine similarities for examples in EntailmentBank T2 and ENWN. All three encoders show little overlap between Random and Gold, showing that these embeddings support Deductive Additivity and the condition in Equation~\ref{eq:additive_reasoning_def}. However, the overlap with Partial is substantially higher.
    }
    \label{fig:embdist}
\end{figure}

\section{Experiments}
\subsection{Intrinsic Evaluation}

We perform two intrinsic evaluations to test if encoders exhibit the deductive additivity property: do they rank gold premise pairs in the proof generation task above incorrect pairs?

\paragraph{Comparing Deduction Embedding Representations}

In our first intrinsic evaluation, we measure the cosine similarity distributions of premise pairs and a deduction in three settings to test for deductive additivity. The first setting uses a deduction $d_{ab}$ and measures the cosine similarity of its embedding $E(d_{ab})$ with a random premise pair $P_r = \{p_x, p_y\}$ where $p_x$ and $p_y$ are drawn randomly from the set of premises, $U(P)$.  The next setting looks at partially random premise pairs, $P_{p} = \{p_a, p_y\}$ where $p_a$ is one of the gold premises $P_g = \{p_a, p_b\}$ that yield the deduction $d_{ab}$. Finally, we measure the distribution of scores for the gold premise pair $P_{g}$ and the following deduction from those premises $d_{ab}$. These three settings correspond to \textbf{Random}, \textbf{Partial}, and \textbf{Gold}, respectively, in Figure~\ref{fig:embdist}.

Additionally, we also compared the gold premise pair $P_g = \{p_a, p_b\}$ with model-generated deductions $S_d(p_a, p_b) = d'_{ab}$ and measured their cosine similarity $cos(\mathbf{e}^{'}_{a+b}, E(d'_{ab}))$.  Finally, we measured the cosine similarity scores of the annotated deductions and the generated deductions $\cos(E(d_{ab}), E(d'_{ab}))$; this is a sort of sanity check to see if the deductive additivity property holds for proof generation. This experiment checks whether the step model introduces significant deviation in embedding similarity compared to using the gold steps. These settings correspond to \textbf{Model} and \textbf{G. to S.} respectively in Figure~\ref{fig:embdist}, all settings have their averages reported in Table~\ref{tab:embedding_reconstruction} in Section~\ref{sec:embedding_reconstruction} of the Appendix as well.

\paragraph{Embedding Representations Results}

Figure~\ref{fig:embdist} shows a slight overlap between the cosine similarity score distributions of random and gold pairs, aligning with expectations and showing that Equation~\ref{eq:additive_reasoning_def} roughly holds for all three encoders. However, the partial pairs have much more overlap with the distribution of gold pairs for each encoder. Concerningly, the partial pairs are much more numerous because these pair one of the ground truth statements with an irrelevant statement, forming a pair we do not want the heuristic to surface. We will see the performance ramifications of this in the end-to-end evaluation. On a positive note, we also see high agreement between the gold premise pair and the generated deduction, indicating that deductions generated by the step model are similar to the annotated deductions.

\begin{table}[h!]
    \centering
    \small
    \begin{tabular}{r c c c     }
        \toprule
         & \multicolumn{2}{c}{\textbf{EB T2}} & \multicolumn{1}{c}{\textbf{ENWN}}\\
        \textbf{Heuristic} & \textbf{Deductive} & \textbf{Goal}  &\textbf{Deductive}   \\
        \midrule
        BM25 &0.47&0.21&0.50\\
        SCSearch &\textbf{0.78}&\textbf{0.39}&\textbf{0.82}\\
        SimCSE (DA) &0.46&0.20&0.59\\
        GPT3-tuned (DA) & 0.54 & 0.23 & 0.54\\
        GPT3 (DA) &0.54&0.24&0.56\\
        \bottomrule
    \end{tabular}
    \caption{
    Comparison against different heuristics on the MRR of selecting gold premises conditioned on their immediate deduction and the goal of the tree. GPT3 outperforms BM25, indicating that there are more complex reasoning steps required than just lexical overlap. However, SCSearch still outperforms all methods by as much as 0.24.
    }
    \label{tab:benchmarks}
\end{table}

\paragraph{Ranking Gold Steps}

The second intrinsic evaluation measures the rankings of premise pairs, $P_{\mathrm{pairs}}$, conditioned on a deduction embedding, $E(d_{ab})$, where one pair is the gold premise pair $P_g = \{p_a, p_b\}$ which yield the deduction. All other pairs are either random $P_r = \{p_x, p_y\}$, where $p_x$ and $p_y$ are sampled uniformly from the set of premises $U(P)$, or are partially random $P_p = \{p_a, p_y\}$.  The full list of premise pairs is the union of all these sets $P_{\mathrm{pairs}} = P_g \cup P_p \cup P_r$.  
We calculate scores for each pair according to how each heuristic scores premise pairs, $\mathrm{scores} = \{\mathrm{heuristic}(P_s, d_{ab}) \mid P_s \in P_{\mathrm{pairs}}\}$. For the heuristics using deductive additivity (DA), the scores are cosine similarities, $\mathrm{scores} = \{\cos(\mathbf{e}^{'}_{n+m}, E(d_{ab})) \mid \{p_n, p_m\} \in P_{\mathrm{pairs}}\}$.  Finally, we sort $\mathrm{scores}$ and find the rank of the gold premise pair. 

We calculate the mean reciprocal rank (MRR) using the ranks of the gold premise pairs across all examples in the EntailmentBank T2 and Everyday Norms: Why Not datasets. We also repeat this process for EntailmentBank T2 where we make the target of the search the claim $g$ instead of the immediate deduction $d_{ab}$. Because the claim $g$ is often a product of multiple deductions in the premise set $P$, we expect the MRR scores to be lower than the scores on the immediate deductions $d_{ab}$. ENWN does not have a T2 setting, so we do not show the claim-conditioned scores because every premise would be related to the claim $g$, making nearly all pairs valid. These are shown in Table~\ref{tab:benchmarks}. A number closer to $1.0$ indicates that the gold premise pair was consistently ranked higher than partial and random premise pairs.

\paragraph{Gold Steps MRR Results}

Table~\ref{tab:benchmarks} shows the BM25 MRR scores as being quite competitive with the methods using deductive additivity, SimCSE, GPT3, and GPT3-tuned, all of which are within $0.1$ of each other. BM25s high performance indicates that the datasets EB T2 and ENWN have many examples where the lexical overlap is enough to determine the gold premise pair $P_g$. GPT3 does outperform the BM25 baseline, however, and in nearly every case, the SimCSE heuristic does as well (except for ENWN). GPT3-tuned does slightly worse in both EB T2 and ENWN, showing that fine-tuning the embeddings to produce the deductive additivity property is not trivial. The degradation in performance is surprising given that the model was fine-tuned on a task very similar to the intrinsic evaluation being reported in Table~\ref{tab:benchmarks}. SCSearch still outperforms all leading methods. There is a significant drop across all methods between ranking premise pairs with the immediate deduction and the goal. Although this was expected, the drop is quite significant and is worth exploring further in future work on how it could be mitigated.

\subsection{Extrinsic Evaluation: Generating Proofs}

 Next, we explore how well heuristics employing deductive additivity can perform on proof generation datasets detailed in Section~\ref{sec:datasets}. 

\begin{table}[t!]
    \centering
    \small
    \begin{tabular}{r c c c c }
        \toprule
        & \multicolumn{2}{c}{\textbf{EB}} & \multicolumn{2}{c}{\textbf{ENWN}} \\
        &  \textbf{Solved} & \textbf{Steps} & \textbf{Solved} & \textbf{Steps} \\
        \midrule
        BM25 & 43\% & 2.2 & 48\% & 5.1 \\
        SCSearch & \textbf{61\%} & 3.4 & \textbf{86\%} & 9.8 \\
        SimCSE (DA) & 44\% & 2.8 & 46\% & 2.7 \\
        GPT3-tuned (DA) & 49\% & 2.1 & 46\% & 2.5\\
        GPT3 (DA) & 49\% & 2.2 & 41\% & 2.2 \\
        \bottomrule
    \end{tabular}
    \caption{
Generated proofs per heuristic on the two datasets.  BM25 has high performance on both of these datasets, indicating that textual overlap is enough to plan reasoning steps for nearly 50\% of the examples.  SimCSE (DA) and GPT3 (DA) underperform BM25 on ENWN; this could mean that these methods are not as sensitive to lexical overlap as BM25 is.  SCSearch still outperforms every baseline by as much as 38\%, showing that a lot of reasoning is unaccounted for in the other methods.
    }
    \label{tab:main_exps}
\end{table}

\paragraph{Results}

We report the percentage of proofs that entailed the goal, $g$, as well as the average number of steps to prove the claim across all planning heuristics in Table~\ref{tab:main_exps}. GPT3 (DA), GPT3-tuned (DA), and SimCSE (DA) are all able to produce slightly more proofs than BM25 on the EB T2 dataset but fail to outperform BM25 on ENWN. Because BM25 is a limited heuristic that only employs lexical overlap, this result shows that nearly 50\% of examples in these datasets can have proofs generated using simple heuristics that use no deeper semantic representations. However, deeper reasoning does help, as shown by the fact that SCSearch is able to generate far more proofs than the other methods across both datasets by as much as 36\%. This finding is also supported by the MRR results of the second intrinsic evaluation, shown in Table~\ref{tab:benchmarks}. Disappointingly, deductive additivity does not seem to be able to capture the same sort of benefits in the heuristic it provides.

\subsection{Single-Step Reasoning Contrast Dataset}

To best understand where the vector-based methods are lacking in performance and pinpoint where improvements can be made, we test each method across a variety of types of reasoning and common failure cases in the Single-Step Reasoning Contrast (SSRC) dataset. In this experiment, we perform the same evaluation as our second intrinsic evaluation, Ranking Gold Steps. Here we use examples from the SSRC dataset, which have been curated and labeled to allow for a report of an MRR on different types of deductions and error cases.

\paragraph{Results}
\begin{figure}[t!]
    \centering
    
    \includegraphics[width=1.0\linewidth]{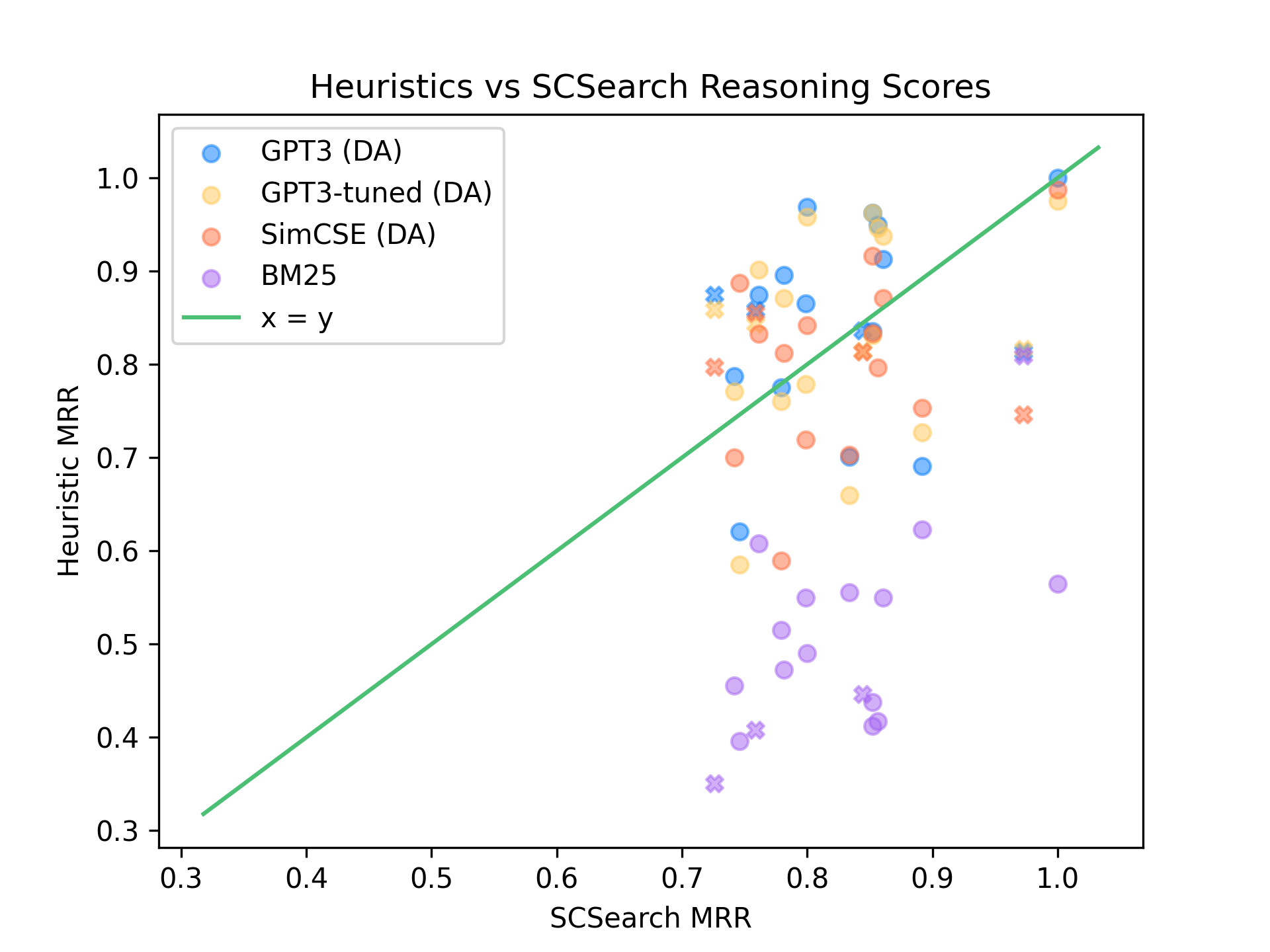}
   
    \caption{ Comparison plot of the heuristic methods versus the SCSearch heuristic.  If a point is above the green line, then that method outperformed SCSearch.  Circles indicate reasoning categories, and X-marks indicate perturbation types.  BM25 underperforms all other methods, showing that the dataset is not sensitive to lexical overlap. }
    \label{fig:reasoning_scatter}
\end{figure}

\begin{table}[h!]
    \centering
    \small
    \begin{tabular}{r c  }
        \toprule
        Encoder &  \textbf{Overall MRR}  \\
        \midrule
        SCSearch & 0.83 \\
        SimCSE (DA) & 0.80 \\
        GPT3-Tuned (DA) & 0.83 \\
        \midrule GPT3 (DA) & \textbf{0.85} \\
        BM25 & 0.50 \\
        \bottomrule
    \end{tabular}
    \caption{
        Overall scores of each heuristic on the SSRC dataset. GPT3 (AD) outperforms SCSearch slightly on this benchmark, slightly contradicting the results of the previous experiments.
    }
    \label{tab:overall_scores}
\end{table}

Table~\ref{tab:overall_scores} shows the averaged MRR scores across all methods. GPT3 (DA) outperforms SCSearch slightly overall, but to better understand the performance, we plot the average MRR across the fourteen reasoning categories and perturbation types for each method compared to SCSearch in Figure~\ref{fig:reasoning_scatter}. GPT3 (DA) can outperform both BM25 and SimCSE (DA) consistently across nearly every reasoning category and all perturbation types. Furthermore, we see that GPT3 (DA) is capable of beating or matching SCSearch on half of the reasoning categories and perturbation types, contradicting previous results indicating that these datasets might be skewed in areas where SCSearch excels at. 

GPT3-Tuned (DA) performs worse in 9 categories than GPT3 (DA) and better in only 3. This could be from the skewed reasoning categories in EntailmentBank, but it could also be that enforcing the condition in Equation~\ref{eq:additive_reasoning_def} directly is counterproductive. Averaged scores for each reasoning category and perturbation type can be found in Appendix~\ref{sec:ssrc_results}, in Tables \ref{tab:rb_reasoning_class_tbl} and \ref{tab:rb_perturbation_tbl} respectively.

\section{Discussion}

\paragraph{Vector-based methods are not sufficient to capture all information for planning deductions.}
We've found that vector-based methods can represent complex reasoning but fall short in planning reasoning steps when compared to early-fusion premise rankers like SCSearch. Our results suggest a more complex 
 and structured approaches may be necessary for step-by-step systems.

\paragraph{Skewed datasets provide optimistic benchmarks for weaker models.}
Our results focused on the T2 setting because we discovered that a BM25 + SCSearch pipeline did quite well and scaled to large numbers of premises. However, we believe this is an optimistic result and may not scale to production settings where claims may require more complex deductions that are less sensitive to lexical overlap. Developing datasets with more complex reasoning and benchmarking in real production settings is a focus for future work.

\paragraph{Training for Deductive Additivity can harm performance.}
We found that training deductive additivity directly improves categories of reasoning prevalent in the training dataset while harming other categories. Both larger and more diverse datasets may be a solution for this problem, but GPT3 embeddings already show deductive additivity without explicitly training for it. Developing different training objectives that result in embeddings with deductive additivity is another focus for future work.

\section{Related work}

Our work follows from models and methods done in the Question Answering domain where models are required to generate an answer or select evidence that leads to the answer through ``multi-hop'' reasoning \citep{chen2021multihop, min-etal-2019-multi, nishida-etal-2019-answering}.  Although these end-to-end methods can be used in proof generation, understanding the underlying reasoning of the decisions being made is impactful for understanding the affordances of the model \citep{hase-bansal-2020-evaluating,BansalEtAl2021}. 

Step-by-step methods have been looked at for proof generation, detangling planning and reasoning into separate subsystems that work together as a whole when proving a claim  \citep{dalvi-etal-2021-explaining, ribeiroIRGR, bostrom-etal-2022-natural, yang-etal-2022-generating, hong-etal-2022-metgen, creswell2023selectioninference, yang2023metaqnl}.  There has also been work on using similar modular systems in answering questions with a knowledge base and different types of embeddings \cite{NIPS2013_1cecc7a7, ren2020query2box, tran2022comparative}. Our work extends from this literature, focusing on exploring alternative heuristics for natural language deduction planning entirely in embedding space by tapping into the property of deductive additivity.

We also follow work being done in retrieval, which focuses on finding evidence from a large corpus that would help answer a query. State-of-the-art retrieval methods involve encoding the corpus into vector indexes that can be used to calculate the cosine similarity of an encoded query \citep{xiong2021answering, karpukhin-etal-2020-dense, colbert}. Sparse encoders, like BM25, have also been used to help reduce the search space for relevant passages \citep{valentino2022hybrid}. However, none of the methods tap into the deductive additivity property in their embedding spaces and instead encode the query to find relevant passages and then re-encode the query with the appended passages to find additional relevant passages. We consider this to be similar to early-fusion premise rankers in the proof generation task. 

Another line of relevant work deals with understanding reasoning errors from language models, like the detection of logical fallacies in text \citep{jin-etal-2022-logical}. We further this line of work with the SSRC dataset, building a contrast set \citep{gardner-etal-2020-evaluating} for reasoning targeting certain types of deductions and common reasoning errors.

\section{Conclusion}

In this work, we have explored the property of deductive additivity in sentence embedding spaces. Results show that off-the-shelf sentence encoders exhibit the property somewhat; however, when used as heuristics in natural language proof generation, they are only slightly more successful than BM25. Furthermore, we see that fine-tuning for deductive additivity does not lead to better reasoning capabilities of the embedding space, and we posit that a large contributor to this could be skewed datasets. We introduced the Single-Step Reasoning Contrast dataset, which shows that these same skewed datasets provide over-optimistic results for inferior methods harming our ability to benchmark systems for their use in production settings. Lastly, we've shown that early-fusion premise rankers like SCSearch still outperform vector-based approaches. However, their ability to scale to more diverse reasoning datasets that are less sensitive to lexical overlap is still an open question for future work.


\section*{Acknowledgments}

This work was partially supported by NSF CAREER Award IIS-2145280, the NSF AI Institute for Foundations of Machine Learning (IFML), a gift from Salesforce, Inc., a gift from Adobe, and a grant from Open Philanthropy. Thanks to the anonymous reviewers for their helpful comments.


\bibliography{custom}

\begin{thebibliography}{36}
\expandafter\ifx\csname natexlab\endcsname\relax\def\natexlab#1{#1}\fi

\bibitem[{Bansal et~al.(2021)Bansal, Wu, Zhou, Fok, Nushi, Kamar, Ribeiro, and
  Weld}]{BansalEtAl2021}
Gagan Bansal, Tongshuang Wu, Joyce Zhou, Raymond Fok, Besmira Nushi, Ece Kamar,
  Marco~Tulio Ribeiro, and Daniel Weld. 2021.
\newblock \href {https://doi.org/10.1145/3411764.3445717} {{Does the Whole
  Exceed Its Parts? The Effect of AI Explanations on Complementary Team
  Performance}}.
\newblock In \emph{Proceedings of the 2021 CHI Conference on Human Factors in
  Computing Systems}, New York, NY, USA. Association for Computing Machinery.

\bibitem[{Bordes et~al.(2013)Bordes, Usunier, Garcia-Duran, Weston, and
  Yakhnenko}]{NIPS2013_1cecc7a7}
Antoine Bordes, Nicolas Usunier, Alberto Garcia-Duran, Jason Weston, and Oksana
  Yakhnenko. 2013.
\newblock \href
  {https://proceedings.neurips.cc/paper_files/paper/2013/file/1cecc7a77928ca8133fa24680a88d2f9-Paper.pdf}
  {Translating embeddings for modeling multi-relational data}.
\newblock In \emph{Advances in Neural Information Processing Systems},
  volume~26. Curran Associates, Inc.

\bibitem[{Bostrom et~al.(2022)Bostrom, Sprague, Chaudhuri, and
  Durrett}]{bostrom-etal-2022-natural}
Kaj Bostrom, Zayne Sprague, Swarat Chaudhuri, and Greg Durrett. 2022.
\newblock \href {https://aclanthology.org/2022.findings-emnlp.358} {Natural
  language deduction through search over statement compositions}.
\newblock In \emph{Findings of the Association for Computational Linguistics:
  EMNLP 2022}, pages 4871--4883, Abu Dhabi, United Arab Emirates. Association
  for Computational Linguistics.

\bibitem[{Bostrom et~al.(2021)Bostrom, Zhao, Chaudhuri, and
  Durrett}]{bostrom-etal-2021-flexible}
Kaj Bostrom, Xinyu Zhao, Swarat Chaudhuri, and Greg Durrett. 2021.
\newblock \href {https://doi.org/10.18653/v1/2021.emnlp-main.506} {Flexible
  generation of natural language deductions}.
\newblock In \emph{Proceedings of the 2021 Conference on Empirical Methods in
  Natural Language Processing}, pages 6266--6278, Online and Punta Cana,
  Dominican Republic. Association for Computational Linguistics.

\bibitem[{Brown et~al.(2020)Brown, Mann, Ryder, Subbiah, Kaplan, Dhariwal,
  Neelakantan, Shyam, Sastry, Askell, Agarwal, Herbert-Voss, Krueger, Henighan,
  Child, Ramesh, Ziegler, Wu, Winter, Hesse, Chen, Sigler, Litwin, Gray, Chess,
  Clark, Berner, McCandlish, Radford, Sutskever, and
  Amodei}]{NEURIPS2020_1457c0d6}
Tom Brown, Benjamin Mann, Nick Ryder, Melanie Subbiah, Jared~D Kaplan, Prafulla
  Dhariwal, Arvind Neelakantan, Pranav Shyam, Girish Sastry, Amanda Askell,
  Sandhini Agarwal, Ariel Herbert-Voss, Gretchen Krueger, Tom Henighan, Rewon
  Child, Aditya Ramesh, Daniel Ziegler, Jeffrey Wu, Clemens Winter, Chris
  Hesse, Mark Chen, Eric Sigler, Mateusz Litwin, Scott Gray, Benjamin Chess,
  Jack Clark, Christopher Berner, Sam McCandlish, Alec Radford, Ilya Sutskever,
  and Dario Amodei. 2020.
\newblock \href
  {https://proceedings.neurips.cc/paper_files/paper/2020/file/1457c0d6bfcb4967418bfb8ac142f64a-Paper.pdf}
  {Language models are few-shot learners}.
\newblock In \emph{Advances in Neural Information Processing Systems},
  volume~33, pages 1877--1901. Curran Associates, Inc.

\bibitem[{Chen et~al.(2019)Chen, Lin, and Durrett}]{chen2021multihop}
Jifan Chen, Shih{-}ting Lin, and Greg Durrett. 2019.
\newblock \href {https://arxiv.org/abs/1910.02610} {Multi-hop question
  answering via reasoning chains}.
\newblock \emph{arXiv}, abs/1910.02610.

\bibitem[{Chen et~al.(2020{\natexlab{a}})Chen, Kornblith, Norouzi, and
  Hinton}]{pmlr-v119-chen20j}
Ting Chen, Simon Kornblith, Mohammad Norouzi, and Geoffrey Hinton.
  2020{\natexlab{a}}.
\newblock \href {https://proceedings.mlr.press/v119/chen20j.html} {A simple
  framework for contrastive learning of visual representations}.
\newblock In \emph{Proceedings of the 37th International Conference on Machine
  Learning}, volume 119 of \emph{Proceedings of Machine Learning Research},
  pages 1597--1607. PMLR.

\bibitem[{Chen et~al.(2020{\natexlab{b}})Chen, Fan, Girshick, and
  He}]{DBLP:journals/corr/abs-2003-04297}
Xinlei Chen, Haoqi Fan, Ross~B. Girshick, and Kaiming He. 2020{\natexlab{b}}.
\newblock \href {http://arxiv.org/abs/2003.04297} {Improved baselines with
  momentum contrastive learning}.
\newblock \emph{CoRR}, abs/2003.04297.

\bibitem[{Creswell et~al.(2023)Creswell, Shanahan, and
  Higgins}]{creswell2023selectioninference}
Antonia Creswell, Murray Shanahan, and Irina Higgins. 2023.
\newblock \href {https://openreview.net/forum?id=3Pf3Wg6o-A4}
  {Selection-inference: Exploiting large language models for interpretable
  logical reasoning}.
\newblock In \emph{The Eleventh International Conference on Learning
  Representations}.

\bibitem[{Dalvi et~al.(2021)Dalvi, Jansen, Tafjord, Xie, Smith, Pipatanangkura,
  and Clark}]{dalvi-etal-2021-explaining}
Bhavana Dalvi, Peter Jansen, Oyvind Tafjord, Zhengnan Xie, Hannah Smith,
  Leighanna Pipatanangkura, and Peter Clark. 2021.
\newblock \href {https://doi.org/10.18653/v1/2021.emnlp-main.585} {Explaining
  answers with entailment trees}.
\newblock In \emph{Proceedings of the 2021 Conference on Empirical Methods in
  Natural Language Processing}, pages 7358--7370, Online and Punta Cana,
  Dominican Republic. Association for Computational Linguistics.

\bibitem[{Gao et~al.(2021)Gao, Yao, and Chen}]{gao-etal-2021-simcse}
Tianyu Gao, Xingcheng Yao, and Danqi Chen. 2021.
\newblock \href {https://doi.org/10.18653/v1/2021.emnlp-main.552} {{S}im{CSE}:
  Simple contrastive learning of sentence embeddings}.
\newblock In \emph{Proceedings of the 2021 Conference on Empirical Methods in
  Natural Language Processing}, pages 6894--6910, Online and Punta Cana,
  Dominican Republic. Association for Computational Linguistics.

\bibitem[{Gardner et~al.(2020)Gardner, Artzi, Basmov, Berant, Bogin, Chen,
  Dasigi, Dua, Elazar, Gottumukkala, Gupta, Hajishirzi, Ilharco, Khashabi, Lin,
  Liu, Liu, Mulcaire, Ning, Singh, Smith, Subramanian, Tsarfaty, Wallace,
  Zhang, and Zhou}]{gardner-etal-2020-evaluating}
Matt Gardner, Yoav Artzi, Victoria Basmov, Jonathan Berant, Ben Bogin, Sihao
  Chen, Pradeep Dasigi, Dheeru Dua, Yanai Elazar, Ananth Gottumukkala, Nitish
  Gupta, Hannaneh Hajishirzi, Gabriel Ilharco, Daniel Khashabi, Kevin Lin,
  Jiangming Liu, Nelson~F. Liu, Phoebe Mulcaire, Qiang Ning, Sameer Singh,
  Noah~A. Smith, Sanjay Subramanian, Reut Tsarfaty, Eric Wallace, Ally Zhang,
  and Ben Zhou. 2020.
\newblock \href {https://doi.org/10.18653/v1/2020.findings-emnlp.117}
  {Evaluating models{'} local decision boundaries via contrast sets}.
\newblock In \emph{Findings of the Association for Computational Linguistics:
  EMNLP 2020}, pages 1307--1323, Online. Association for Computational
  Linguistics.

\bibitem[{Hase and Bansal(2020)}]{hase-bansal-2020-evaluating}
Peter Hase and Mohit Bansal. 2020.
\newblock \href {https://doi.org/10.18653/v1/2020.acl-main.491} {Evaluating
  explainable {AI}: Which algorithmic explanations help users predict model
  behavior?}
\newblock In \emph{Proceedings of the 58th Annual Meeting of the Association
  for Computational Linguistics}, pages 5540--5552, Online. Association for
  Computational Linguistics.

\bibitem[{He et~al.(2020)He, Fan, Wu, Xie, and Girshick}]{he2020momentum}
Kaiming He, Haoqi Fan, Yuxin Wu, Saining Xie, and Ross Girshick. 2020.
\newblock Momentum contrast for unsupervised visual representation learning.
\newblock In \emph{Proceedings of the IEEE/CVF conference on computer vision
  and pattern recognition}, pages 9729--9738.

\bibitem[{Hong et~al.(2022)Hong, Zhang, Yu, and Zhang}]{hong-etal-2022-metgen}
Ruixin Hong, Hongming Zhang, Xintong Yu, and Changshui Zhang. 2022.
\newblock \href {https://doi.org/10.18653/v1/2022.findings-naacl.145}
  {{METGEN}: A module-based entailment tree generation framework for answer
  explanation}.
\newblock In \emph{Findings of the Association for Computational Linguistics:
  NAACL 2022}, pages 1887--1905, Seattle, United States. Association for
  Computational Linguistics.

\bibitem[{Jin et~al.(2022)Jin, Lalwani, Vaidhya, Shen, Ding, Lyu, Sachan,
  Mihalcea, and Schoelkopf}]{jin-etal-2022-logical}
Zhijing Jin, Abhinav Lalwani, Tejas Vaidhya, Xiaoyu Shen, Yiwen Ding, Zhiheng
  Lyu, Mrinmaya Sachan, Rada Mihalcea, and Bernhard Schoelkopf. 2022.
\newblock \href {https://aclanthology.org/2022.findings-emnlp.532} {Logical
  fallacy detection}.
\newblock In \emph{Findings of the Association for Computational Linguistics:
  EMNLP 2022}, pages 7180--7198, Abu Dhabi, United Arab Emirates. Association
  for Computational Linguistics.

\bibitem[{Karpukhin et~al.(2020)Karpukhin, Oguz, Min, Lewis, Wu, Edunov, Chen,
  and Yih}]{karpukhin-etal-2020-dense}
Vladimir Karpukhin, Barlas Oguz, Sewon Min, Patrick Lewis, Ledell Wu, Sergey
  Edunov, Danqi Chen, and Wen-tau Yih. 2020.
\newblock \href {https://doi.org/10.18653/v1/2020.emnlp-main.550} {Dense
  passage retrieval for open-domain question answering}.
\newblock In \emph{Proceedings of the 2020 Conference on Empirical Methods in
  Natural Language Processing (EMNLP)}, pages 6769--6781, Online. Association
  for Computational Linguistics.

\bibitem[{Khattab and Zaharia(2020)}]{colbert}
Omar Khattab and Matei Zaharia. 2020.
\newblock \href {https://doi.org/10.1145/3397271.3401075} {Colbert: Efficient
  and effective passage search via contextualized late interaction over bert}.
\newblock In \emph{Proceedings of the 43rd International ACM SIGIR Conference
  on Research and Development in Information Retrieval}, SIGIR '20, page
  39–48, New York, NY, USA. Association for Computing Machinery.

\bibitem[{Liu et~al.(2022)Liu, Swayamdipta, Smith, and
  Choi}]{liu-etal-2022-wanli}
Alisa Liu, Swabha Swayamdipta, Noah~A. Smith, and Yejin Choi. 2022.
\newblock \href {https://aclanthology.org/2022.findings-emnlp.508} {{WANLI}:
  Worker and {AI} collaboration for natural language inference dataset
  creation}.
\newblock In \emph{Findings of the Association for Computational Linguistics:
  EMNLP 2022}, pages 6826--6847, Abu Dhabi, United Arab Emirates. Association
  for Computational Linguistics.

\bibitem[{Min et~al.(2019)Min, Zhong, Zettlemoyer, and
  Hajishirzi}]{min-etal-2019-multi}
Sewon Min, Victor Zhong, Luke Zettlemoyer, and Hannaneh Hajishirzi. 2019.
\newblock \href {https://doi.org/10.18653/v1/P19-1613} {Multi-hop reading
  comprehension through question decomposition and rescoring}.
\newblock In \emph{Proceedings of the 57th Annual Meeting of the Association
  for Computational Linguistics}, pages 6097--6109, Florence, Italy.
  Association for Computational Linguistics.

\bibitem[{Musgrave et~al.(2020)Musgrave, Belongie, and
  Lim}]{Musgrave2020PyTorchML}
Kevin Musgrave, Serge~J. Belongie, and Ser-Nam Lim. 2020.
\newblock {PyTorch Metric Learning}.
\newblock \emph{ArXiv}, abs/2008.09164.

\bibitem[{Nishida et~al.(2019)Nishida, Nishida, Nagata, Otsuka, Saito, Asano,
  and Tomita}]{nishida-etal-2019-answering}
Kosuke Nishida, Kyosuke Nishida, Masaaki Nagata, Atsushi Otsuka, Itsumi Saito,
  Hisako Asano, and Junji Tomita. 2019.
\newblock \href {https://doi.org/10.18653/v1/P19-1225} {Answering while
  summarizing: Multi-task learning for multi-hop {QA} with evidence
  extraction}.
\newblock In \emph{Proceedings of the 57th Annual Meeting of the Association
  for Computational Linguistics}, pages 2335--2345, Florence, Italy.
  Association for Computational Linguistics.

\bibitem[{OpenAI(2023)}]{openai2023gpt4}
OpenAI. 2023.
\newblock \href {http://arxiv.org/abs/2303.08774} {{GPT-4 Technical Report}}.

\bibitem[{Radford et~al.(2021)Radford, Kim, Hallacy, Ramesh, Goh, Agarwal,
  Sastry, Askell, Mishkin, Clark et~al.}]{radford2021learning}
Alec Radford, Jong~Wook Kim, Chris Hallacy, Aditya Ramesh, Gabriel Goh,
  Sandhini Agarwal, Girish Sastry, Amanda Askell, Pamela Mishkin, Jack Clark,
  et~al. 2021.
\newblock Learning transferable visual models from natural language
  supervision.
\newblock In \emph{International conference on machine learning}, pages
  8748--8763. PMLR.

\bibitem[{Ren et~al.(2020)Ren, Hu, and Leskovec}]{ren2020query2box}
Hongyu Ren, Weihua Hu, and Jure Leskovec. 2020.
\newblock \href {https://openreview.net/forum?id=BJgr4kSFDS} {Query2box:
  Reasoning over knowledge graphs in vector space using box embeddings}.
\newblock In \emph{International Conference on Learning Representations}.

\bibitem[{Ribeiro et~al.(2022)Ribeiro, Wang, Ma, Zhu, Dong, Chen, Peng, Huang,
  Arnold, and Roth}]{ribeiroIRGR}
Danilo~Neves Ribeiro, Shen Wang, Xiaofei Ma, Henghui Zhu, Rui Dong, Xinchi
  Chen, Zhu Peng, Zhiheng Huang, Andrew Arnold, and Dan Roth. 2022.
\newblock \href {https://arxiv.org/abs/2205.09224} {Entailment tree
  explanations via iterative retrieval-generation reasoner}.
\newblock In \emph{Proceedings of the 2022 Conference of the North American
  Chapter of the Association for Computational Linguistics: Human Language
  Technologies}, Online and Seattle, USA. Association for Computational
  Linguistics.

\bibitem[{Robertson et~al.(1995)Robertson, Walker, Jones, Hancock-Beaulieu, and
  Gatford}]{robertson1995okapi}
Stephen Robertson, S.~Walker, S.~Jones, M.~M. Hancock-Beaulieu, and M.~Gatford.
  1995.
\newblock \href
  {https://www.microsoft.com/en-us/research/publication/okapi-at-trec-3/}
  {{Okapi at TREC-3}}.
\newblock In \emph{Overview of the Third Text REtrieval Conference (TREC-3)},
  pages 109--126. Gaithersburg, MD: NIST.

\bibitem[{Sprague et~al.(2022)Sprague, Bostrom, Chaudhuri, and
  Durrett}]{sprague-etal-2022-natural}
Zayne Sprague, Kaj Bostrom, Swarat Chaudhuri, and Greg Durrett. 2022.
\newblock \href {https://aclanthology.org/2022.emnlp-main.564} {Natural
  language deduction with incomplete information}.
\newblock In \emph{Proceedings of the 2022 Conference on Empirical Methods in
  Natural Language Processing}, pages 8230--8258, Abu Dhabi, United Arab
  Emirates. Association for Computational Linguistics.

\bibitem[{Tran et~al.(2022)Tran, Phan, Quach, Nguyen, Jo, and
  Nguyen}]{tran2022comparative}
Khiem~Vinh Tran, Hao~Phu Phan, Khang Nguyen~Duc Quach, Ngan Luu-Thuy Nguyen,
  Jun Jo, and Thanh~Tam Nguyen. 2022.
\newblock A comparative study of question answering over knowledge bases.
\newblock In \emph{International Conference on Advanced Data Mining and
  Applications}, pages 259--274. Springer.

\bibitem[{Valentino et~al.(2022)Valentino, Thayaparan, Ferreira, and
  Freitas}]{valentino2022hybrid}
Marco Valentino, Mokanarangan Thayaparan, Deborah Ferreira, and Andr{\'e}
  Freitas. 2022.
\newblock Hybrid autoregressive inference for scalable multi-hop explanation
  regeneration.
\newblock In \emph{Proceedings of the AAAI Conference on Artificial
  Intelligence}, volume~36, pages 11403--11411.

\bibitem[{Xiong et~al.(2021)Xiong, Li, Iyer, Du, Lewis, Wang, Mehdad, Yih,
  Riedel, Kiela, and Oguz}]{xiong2021answering}
Wenhan Xiong, Xiang Li, Srini Iyer, Jingfei Du, Patrick Lewis, William~Yang
  Wang, Yashar Mehdad, Scott Yih, Sebastian Riedel, Douwe Kiela, and Barlas
  Oguz. 2021.
\newblock \href {https://openreview.net/forum?id=EMHoBG0avc1} {Answering
  complex open-domain questions with multi-hop dense retrieval}.
\newblock In \emph{International Conference on Learning Representations}.

\bibitem[{Xue et~al.(2023)Xue, Wang, Wang, Han, Yu, and Ji}]{Xue2023RCOTDA}
Tianci Xue, Ziqi Wang, Zhenhailong Wang, Chi Han, Pengfei Yu, and Heng Ji.
  2023.
\newblock {RCOT: Detecting and Rectifying Factual Inconsistency in Reasoning by
  Reversing Chain-of-Thought}.
\newblock \emph{ArXiv}, abs/2305.11499.

\bibitem[{Yang and Deng(2023)}]{yang2023metaqnl}
Kaiyu Yang and Jia Deng. 2023.
\newblock Learning symbolic rules for reasoning in quasi-natural language.
\newblock \emph{Transactions on Machine Learning Research (TMLR)}.

\bibitem[{Yang et~al.(2022)Yang, Deng, and Chen}]{yang-etal-2022-generating}
Kaiyu Yang, Jia Deng, and Danqi Chen. 2022.
\newblock \href {https://aclanthology.org/2022.emnlp-main.7} {Generating
  natural language proofs with verifier-guided search}.
\newblock In \emph{Proceedings of the 2022 Conference on Empirical Methods in
  Natural Language Processing}, pages 89--105, Abu Dhabi, United Arab Emirates.
  Association for Computational Linguistics.

\bibitem[{Ye and Durrett(2022)}]{ye2022unreliability}
Xi~Ye and Greg Durrett. 2022.
\newblock {The Unreliability of Explanations in Few-shot Prompting for Textual
  Reasoning}.
\newblock In \emph{Advances in Neural Information Processing Systems}.

\bibitem[{Zhang et~al.(2023)Zhang, Press, Merrill, Liu, and
  Smith}]{zhang2023language}
Muru Zhang, Ofir Press, William Merrill, Alisa Liu, and Noah~A Smith. 2023.
\newblock How language model hallucinations can snowball.
\newblock \emph{arXiv preprint arXiv:2305.13534}.

\end{thebibliography}
\bibliographystyle{acl_natbib}
\newpage
\appendix

\section{Embedding Reconstruction Results}
\label{sec:embedding_reconstruction}

Table~\ref{tab:embedding_reconstruction} shows the averaged cosine similarity of the random, partially random, and gold pairs, as well as the cosine similarities for the gold pairs with the step model generations. This provides complementary information to Figure~\ref{fig:embdist}.

\begin{table*}[h!]
    \centering
    \small
    \begin{tabular}{r c c c c c c c c c   }
        \toprule
        & \multicolumn{4}{c}{\textbf{EB}} & \multicolumn{4}{c}{\textbf{ENWN}}  \\
        \textbf{Heuristic} &  \textbf{Rand} & \textbf{Partial} & \textbf{Gold} & \textbf{Model} &  \textbf{Rand} & \textbf{Partial} & \textbf{Gold} & \textbf{Model}\\
        \midrule
        SimCSE & 0.25 & 0.62 & 0.85 & 0.85  & 0.14 & 0.48 & 0.72 & 0.76  \\
        GPT3-tuned & 0.31 & 0.70 & 0.90 & 0.90 & 0.56 & 0.74 & 0.86 & 0.87 \\
        GPT3 & 0.79 & 0.88 & 0.93 & 0.94  & 0.79 & 0.89 & 0.95 & 0.95  \\
        \bottomrule
    \end{tabular}
    \caption{
    We look at the average cosine similarity score of different summed premise pairs with their textual deduction embedding. We see large gaps between a Random set of premises and the Partial/Gold set; however, Partial and Gold are less separated. The Model columns show that there is no loss in representing deductions if the deduction is the gold annotation or from the deduction step model.
    }
    \label{tab:embedding_reconstruction}
\end{table*}

\section{SSRC Dataset Examples}
\label{sec:ssrc_examples}

Table~\ref{tab:rb_examples} shows four examples from the SSRC dataset that have been sampled from different reasoning categories and show different perturbation types for the premises.  

\begin{table*}[h!]
    \centering
    \small
    \begin{tabular}{p{1.25in}p{1.25in}p{1.25in}p{1.25in}}
        \toprule

        \textbf{Category and Perturbation} & \textbf{Premises} & \textbf{Conclusion} & \textbf{Perturbed Premises}\\
        \midrule\\
        Categorical Syllogism, Negation & All cats are animals. Whiskers is a cat. & Whiskers is an Animal. & Some cats are not animals.  Whiskers is not a cat. \\
        \midrule \\
        Causal Reasoning, Irrelevant Fact & High levels of stress cause anxiety. Linda has been under a lot of stress lately. & Linda may develop anxiety. & Anxiety disorders can also manifest as physical symptoms. ... \\
        \midrule \\
        Comparative Reasoning, Incorrect Quantifier & John is stronger than Mary. Mary is stronger than Sue. & John is stronger than Sue. & Some women are stronger than Sue. \\
        \midrule \\
        Temporal Reasoning, False Premise & The store is open for 12 hours. The store opens at 9 AM. & The store closes at 9 PM. & The store is open for 10 hours. \\
        \bottomrule
    \end{tabular}
    \caption{Examples taken from the Single-Step Reasoning Contrast (SSRC) dataset.  The Category and Perturbation column shows which reasoning category is used in the deduction as well as what type of perturbation is applied to the premise.  The perturbed premise is then used to create invalid premise pairs (where one premise could be a gold premise, but the other is perturbed) such that when the two are combined, their deduction does not lead to the conclusion.  There are ten examples per reasoning category, and each example has multiple perturbed premises for each of the four perturbation types.}
    \label{tab:rb_examples}
\end{table*}

\section{SSRC Dataset Results}
\label{sec:ssrc_results}
We report the raw scores for both the reasoning categories and perturbation types in Tables \ref{tab:rb_reasoning_class_tbl} and \ref{tab:rb_perturbation_tbl} respectively. 
\begin{table*}[h!]
    \centering
    \small
    \begin{tabular}{r c c c | c c  }
        \toprule
        &  \textbf{SCSearch} & \textbf{SimCSE (DA)} & \textbf{GPT3-tuned (DA)} & \textbf{BM25} & \textbf{GPT3 (DA)}\\
        \midrule
        Analogy  & 0.86 & 0.80 & \textbf{0.95} &0.42& \textbf{0.95} \\
        Categorical syllogism   & 0.80 & 0.72 & 0.78 & 0.55& \textbf{0.87}\\
        Causal Reasoning  & \textbf{0.78} & 0.59 & 0.76 &0.52 & \textbf{0.78} \\
        Classification   & 0.86 & 0.87 & \textbf{0.94} &0.55 & 0.91 \\
        Comparative Reasoning & 0.85 &0.92 & \textbf{0.96} &0.44& \textbf{0.96} \\
        Composition & 0.75 &\textbf{0.89}& 0.59 &0.40& 0.62 \\
        Definition & 0.80 &0.84 & 0.96 &0.49& \textbf{0.97} \\
        Divisions & \textbf{0.85} &0.83 & 0.83 &0.41& 0.84 \\
        Modus Ponens &\textbf{1.0} &0.99 & 0.98 &0.56& \textbf{1.0} \\
        Modus Tollens & \textbf{0.83} &0.70 & 0.66 &0.56& 0.7 \\
        Propositional Logic & \textbf{0.89} &0.75 & 0.73 &0.62& 0.69 \\
        Quantification Logic & 0.76 &0.83 & \textbf{0.90} &0.61& 0.87 \\
        Spatial Reasoning & 0.78 &0.81& 0.87 &0.47& \textbf{0.90} \\
        Temporal Reasoning & 0.74 &0.70 & 0.77 &0.46& \textbf{0.79} \\ 
        \bottomrule
    \end{tabular}
    \caption{Results of each heuristic on SSRC are broken down by the reasoning category and averaged over the individual perturbation types of each category.  We separate SCSearch and SimCSE (DA) from the BM25 and GPT3 (DA) heuristics as the BM25 and GPT3 (DA) have not been trained on any natural language inference data (with the possibility that GPT3 may have seen some incidental examples of inferences in its pretraining), making them close to zero-shot on this task.  SCSearch and SimCSE (DA) have both been fine-tuned on reasoning datasets (EntailmentBank and NLI, respectively). }
    \label{tab:rb_reasoning_class_tbl}
\end{table*}
\begin{table*}[t!]
    \centering
    \small
    \begin{tabular}{r c c c | c c  }
        \toprule
        & \textbf{SCSearch} & \textbf{SimCSE (DA)} & \textbf{GPT3-tuned (DA)} & \textbf{BM25} & \textbf{GPT3 (DA)}\\
        \midrule
        False Premise & \textbf{0.84} & 0.81 & 0.81 & 0.45 & 0.81\\
        Irrelevant Fact & \textbf{0.97} & 0.75 & 0.82 &  0.81 & 0.87\\
        Incorrect Quantification & 0.76 & \textbf{0.86} & 0.84 & 0.41 & \textbf{0.86} \\
        Negated & 0.73 & 0.80 & 0.86 & 0.35 & \textbf{0.87} \\
        \bottomrule
    \end{tabular}
    \caption{
    Results of each heuristic on SSRC are broken down by the perturbation type and averaged over the individual reasoning categories. We again separate SCSearch and SimCSE (DA) from the BM25 and GPT3 (DA) heuristics.
    }
    \label{tab:rb_perturbation_tbl}
\end{table*}

\section{Proof Generation Modules}
\label{sec:appendix_search_modules}

We outline in more detail the proof generation search algorithm we use in our experiments following work from \citet{sprague-etal-2022-natural} and \citet{bostrom-etal-2022-natural}.

\begin{algorithm}[h!]
\small
\caption{The main search function with a heuristic using the deductive additivity property. Given a set of premises and a goal claim, generate intermediate deductions until the claim is proven true or a termination criterion is met. $E$ is a sentence encoder.}
\begin{algorithmic}
\State {\textbf{Input} A list $X$ of string premises $\mathbf{p}_i$ that will be used to search over to prove a string claim $\mathbf{g}$ }
\State {\textbf{Output} A list of steps taken by the algorithm with their generations}
\State {\textbf{Procedure} $\textsc{Search}(X = \{\mathbf{p}_1,\ \dots\ \mathbf{p}_n\},\ \mathbf{g})$:}
\State $f \gets \{E(\mathbf{p}_i) + E(\mathbf{p}_j)\ |\ \mathbf{p}_i, \mathbf{p}_j\ \in\ X,\ i \neq j \}$
\State $\hat{g} \gets E(\mathbf{g})$
\State $gens \gets \{\}$
\State $maxSteps \in \mathbb{N}$
\State $i \gets 1$
\While{$|f| > 0 \wedge i \leq maxSteps$}

    \State $step \gets \argmax\limits_{x_i \in f} M(x_i, \hat{g})$
    \State $f \gets f\ \setminus\ \{step\}$
    \State $\text{sample}\ \mathbf{y}_i\ \text{from}\ p_S(\mathbf{y}\ |\ step)$
    \If{$\mathbf{y}_i \notin gens$}
        \State $gens \gets gens\ \bigcup\ \{\mathbf{y}_i\}$
        \State $\textbf{yield}\ (step, \mathbf{y}_i)$
        \If{$\text{entails}(\mathbf{y}_i, \mathbf{g})$}
            \Return
        \EndIf
        \State $\begin{aligned} f \gets f  \cup \{ E(\mathbf{y}_i)\ +\ E(\mathbf{x}_j)\ | \mathbf{x}_j \in X \} \end{aligned} $
        \State $ f \gets f \cup \{ E(\mathbf{y}_i)\ +\ E(\mathbf{y}_j)\ |\ \mathbf{y}_j \in gens, 1 \leq j < i \} $
       
        \State $i \gets i + 1$
    \EndIf
    
    
\EndWhile
\end{algorithmic}
\label{alg:ar_search}
\end{algorithm}

\subsection{Deductive Step Model}

The deductive step model is trained using the EntailmentBank dataset following \citet{bostrom-etal-2022-natural}. We transform the annotated entailment trees into individual steps $T_i = (x_1, x_2 \rightarrow c)$ and fine-tune a pre-trained language model to generate the deduction given a set of premises. We do not use data from \citep{bostrom-etal-2021-flexible}.

\subsection{Reasoning Validation}

To ensure that the search space generates well-reasoned deductions, we implement a set of validators that examine both the types of steps being taken and the generations produced by the step models following \citet{sprague-etal-2022-natural}. Firstly, we employ a Consanguinity Threshold step to ensure that the search procedure does not permit steps to consist of the same premise or premises that result in immediate deductions. For instance, if $p_a$ and $p_b$ create the deduction $d_{ab}$, we disallow a new step to be $(p_a, d_{ab})$. This approach effectively promotes diversity in the types of steps being taken. We also enforce that no generation from a step model is an exact duplicate of one of the inputs.

Furthermore, to avoid identifying high-ranking pairs of premises that result in illogical deductions due to hallucination, we devise a new validation method to ensure consistency. The Deduction Agreement validator compares the embedding of the added premises $e_{d'}$ with the embedding of the generated deduction $e_d$. If the cosine similarity falls below a threshold $t_{da}$, the step is filtered out. A running average of all $\cos(e_{d'}, e_d)$ scores for previous deductions is maintained. If a branch in the entailment tree generates too many deductions that have low cosine similarity with their summed premises, it will be filtered out.

\subsection{Entailment Scores}

We employ a DeBERTa model, fine-tuned on the MNLI and WaNLI tasks, to assess the entailment of each generated natural language deduction. If a deduction achieves a score above a predefined threshold, $t_g$, it is considered to have recovered the goal $g$. Once a deduction has successfully recovered the goal, we can trace back the steps used to create that specific deduction, resulting in a minimal proof tree that contains only the essential steps required to prove the goal.

\section{SSRC Prompting}
\label{sec:appendix_ssrc_prompts}

We use ChatGPT to prompt GPT3.5 and create the SSRC dataset.  We followed the same template for all reasoning categories and then used a simple Python script to parse out the examples generated.  Below is an example of how we prompted ChatGPT for the reasoning category Classification.  All prompts are given to ChatGPT one after another.  

\begin{figure*}
\begin{mdframed}[backgroundcolor=gray!20, hidealllines=true]
\centering
\begin{tabular}{p{\linewidth-0.25cm}}

\textbf{Creating the ten examples} \\
Create ten deductions that use classification in the example.  A definition of classification in deductions and an example is below:\\
\\
Definition:\\
Classification involves grouping things based on shared properties or characteristics and drawing conclusions based on these groupings.\\
\\
Example:\\
P1: A dog is a type of animal.\\
P2: A cat is a type of animal.\\
C: Dogs and cats are both animals.\\
\\
Each example should have 2 premises and 1 conclusion.  They must all use classification to perform the deduction.\\
\end{tabular}
\label{fig:my_label}
\end{mdframed}
\caption{Prompt given to ChatGPT that creates the original ten deductions for the specific reasoning category.  The premises given in this step are referred to as the ``gold premises''.}
\end{figure*}

\begin{figure*}
\begin{mdframed}[backgroundcolor=gray!20, hidealllines=true]
\centering
\begin{tabular}{p{\linewidth-0.25cm}}

\textbf{Negating the ten examples} \\
For each of the ten generate negations of each premise and conclusion.  If a premise begins with "All... are x" you should both negate the sentence and remove the word "All" at the beginning.  \\
\\
Just write the negations and put it in the format:\\
\\
P1: Negated premise one \\
P2: Negated premise two \\
C: Negated conclusion \\
\end{tabular}
\label{fig:my_label}
\end{mdframed}
\caption{Once the ten reasoning examples have been generated, we then ask ChatGPT to negate the ten examples' gold premises.  }
\end{figure*}

\begin{figure*}
\begin{mdframed}[backgroundcolor=gray!20, hidealllines=true]
\centering
\begin{tabular}{p{\linewidth-0.25cm}}

\textbf{Creating false premises} \\
For each of the ten generate two false premises total, one for each premise.  The false premise should make the original deduction invalid.  Do NOT generate a false conclusion. Do NOT repeat the valid premises or conclusions.  Only generate the False premises.  \\
\\
Put them in this format:\\
\\
False P1: False premise for premise one\\
False P2: False premise for premise two\\
\end{tabular}
\label{fig:my_label}
\end{mdframed}
\caption{After the negated premises are generated, we ask ChatGPT to create false premises for the ten reasoning examples. False premises are not negated premises. Instead, they should employ some common sense from the model to make a statement false. An example from the SSRC dataset is ``a granny smith is a type of fish.'' which is a false statement. }
\end{figure*}

\begin{figure*}
\begin{mdframed}[backgroundcolor=gray!20, hidealllines=true]
\centering
\begin{tabular}{p{\linewidth-0.25cm}}

\textbf{Generating irrelevant facts: Prompt 1} \\
For each of the ten generate two facts that seem related to the deduction but are in fact irrelevant.  They should not contribute to the deduction at all, but they should be close enough to trick someone.  ONLY GENERATE THE NEW FACTS.\\
\\
Put them in this format:\\
\\
Irrelevant Fact 1: Irrelevant fact for premise one\\
Irrelevant Fact 2: Irrelevant fact for premise two\\
\\
\\
\textbf{Generating irrelevant facts: Prompt 2} \\
Do this again. For each of the ten generate two facts that seem related to the deduction but are in fact irrelevant.  They should not contribute to the deduction at all, but they should be close enough to trick someone.  ONLY GENERATE THE NEW FACTS.\\
\\
Put them in this format:\\
\\
Irrelevant Fact 1: Irrelevant fact for premise one\\
Irrelevant Fact 2: Irrelevant fact for premise two\\
\\
\\
\textbf{Generating irrelevant facts: Prompt 3} \\
Generate one more set of facts. For each of the ten generate two facts that seem related to the deduction but are in fact irrelevant.  They should not contribute to the deduction at all, but they should be close enough to trick someone.  ONLY GENERATE THE NEW FACTS.\\
\\
Put them in this format:\\
\\
Irrelevant Fact 1: Irrelevant fact for premise one\\
Irrelevant Fact 2: Irrelevant fact for premise two\\
\end{tabular}
\label{fig:my_label}
\end{mdframed}
\caption{After the false premises are generated, we ask ChatGPT to create irrelevant facts that are true but not helpful in deducing the original conclusion.  We prompt ChatGPT three times for a set of six irrelevant facts per example.}
\end{figure*}

\begin{figure*}
\begin{mdframed}[backgroundcolor=gray!20, hidealllines=true]
\centering
\begin{tabular}{p{\linewidth-0.25cm}}

\textbf{Generating examples with incorrect quantifiers} \\
Now generate premises from the original set of 10 examples that have incorrect quantifiers that would make the conclusion invalid.  Use things like "All, some, none, etc.". Do not write the conclusion.\\
\\
Put them in the format:\\
Incorrect P1: Incorrect quantifiers for p1\\
Incorrect P2: Incorrect quantifiers for p2\\
\end{tabular}

\label{fig:my_label}
\end{mdframed}
\caption{Finally, we prompt ChatGPT to adjust the quantifier on the original gold premises.}
\end{figure*}

\section{Examples of GPT ranking SSRC premise pairs}

Here we show three examples from the SSRC dataset and place the premise pairs in order of how GPT3 ranked them.  The \textbf{Category} indicates which reasoning category the example belongs to, \textbf{Perturbation} indicates which perturbation type the example is exhibiting, \textbf{Target} is the claim $g$, \textbf{Gold Premises} are the correct premises that yield the claim from a deduction, \textbf{Rank} is the Rank GPT3 gave the gold premises (1 being the best).  We also include all premise pairs and their ranks below the \textbf{Rank} of the gold premises, and we mark the pair \textbf{(G)} for the gold premise pair.

\begin{figure*}
\begin{mdframed}[backgroundcolor=gray!20, hidealllines=true]
\centering
\begin{tabular}{p{\linewidth-0.25cm}}

\textbf{Category}:  spatial reasoning\\
\textbf{Perturbation Type}:  NEGATED\\
\textbf{Target}:  The pharmacy is on the same side of the street as the bank.\\
\textbf{Gold Premises}:  The post office is on the same side of the street as the bank. The pharmacy is next to the post office.\\
\textbf{Rank}:  1\\
\quad\textbf{Rank (G) 1}:  the pharmacy is next to the post office. the post office is on the same side of the street as the bank.\\
\quad\textbf{Rank 2}:  the pharmacy is not next to the post office. the post office is on the same side of the street as the bank.\\
\quad\textbf{Rank 3}:  the pharmacy is next to the post office. the post office is not on the same side of the street as the bank.\\
\quad\textbf{Rank 4}:  the pharmacy is not next to the post office. the post office is not on the same side of the street as the bank.\\
\end{tabular}
\label{fig:ssrc_ranking_examples}
\end{mdframed}
\caption{An example of GPT3 embeddings using deductive additivity correctly ranking a spatial reasoning example with negation from the SSRC dataset.}
\end{figure*}

\begin{figure*}
\begin{mdframed}[backgroundcolor=gray!20, hidealllines=true]
\centering
\begin{tabular}{p{\linewidth-0.25cm}}
\textbf{Category}:  definition\\
\textbf{Perturbation Type}:  FALSE PREMISE\\
\textbf{Target}:  A Granny Smith is a type of fruit.\\
\textbf{Gold Premises}:  An apple is a type of fruit. A Granny Smith is a type of apple.\\
\textbf{Rank}:  1\\
\quad\textbf{Rank (G) 1}:  an apple is a type of fruit. a granny smith is a type of apple.\\
\quad\textbf{Rank 2}:  an apple is a type of fruit. a granny smith is a type of fish.\\
\quad\textbf{Rank 3}:  a granny smith is a type of apple. an apple is a type of vegetable.\\
\quad\textbf{Rank 4}:  a granny smith is a type of fish. an apple is a type of vegetable.\\
\end{tabular}
\label{fig:ssrc_ranking_examples}
\end{mdframed}
\caption{An example of GPT3 embeddings using deductive additivity correctly ranking a definition example with false premises from the SSRC dataset.}
\end{figure*}

\begin{figure*}
\begin{mdframed}[backgroundcolor=gray!20, hidealllines=true]
\centering
\begin{tabular}{p{\linewidth-0.25cm}}
\textbf{Category}:  propositional logic\\
\textbf{Perturbation Type}:  IRRELEVANT FACT\\
\textbf{Target}:  The store is closed.\\
\textbf{Gold Premises}:  If it is Sunday, the store is closed. It is Sunday.\\
\textbf{Rank}:  10\\
\quad\textbf{Rank 1}:  i need to buy groceries. if it is sunday, the store is closed.\\
\quad\textbf{Rank 2}:  i forgot to bring my reusable bag. if it is sunday, the store is closed.\\
\quad\textbf{Rank 3}:  if it is sunday, the store is closed. i prefer to shop on saturdays.\\
\quad\textbf{Rank 4}:  the store is near my house. it is sunday.\\
\quad\textbf{Rank 5}:  it is sunday. the store has a sale.\\
\quad\textbf{Rank 6}:  i need to buy groceries. the store has a sale.\\
\quad\textbf{Rank 7}:  i prefer to shop on saturdays. the store has a sale.\\
\quad\textbf{Rank 8}:  i forgot to bring my reusable bag. the store has a sale.\\
\quad\textbf{Rank 9}:  the store is near my house. i prefer to shop on saturdays.\\
\quad\textbf{Rank (G) 10}:  if it is sunday, the store is closed. it is sunday.\\
\quad\textbf{Rank 11}:  the store is near my house. i need to buy groceries.\\
\quad\textbf{Rank 12}:  i have a coupon for the store. it is sunday.\\
\quad\textbf{Rank 13}:  the store is near my house. i forgot to bring my reusable bag.\\
\quad\textbf{Rank 14}:  i have a coupon for the store. i prefer to shop on saturdays.\\
\quad\textbf{Rank 15}:  i have a coupon for the store. i need to buy groceries.\\
\quad\textbf{Rank 16}:  i have a coupon for the store. i forgot to bring my reusable bag.\\
\end{tabular}
\label{fig:ssrc_ranking_examples}
\end{mdframed}
\caption{An example of GPT3 failing to rank a propositional logic example of the SSRC dataset correctly amongst irrelevant facts.}
\end{figure*}

\end{document}